\documentclass[12pt]{article}
\usepackage{amsmath}
\usepackage{graphicx}
\usepackage{enumerate}
\usepackage{url} 

\usepackage[english]{babel}
\usepackage{setspace} 
\usepackage{hyperref}

\usepackage{amsmath,amssymb,amsfonts}%
\usepackage[table]{xcolor}

\usepackage[backend=biber, bibstyle=apa, citestyle=bwl-FU, natbib=true]{biblatex}
\newcommand{\x}{\mathbf{x}}

\newcommand{\z}{\mathbf{z}}

\newcommand{\R}{\mathbf{R}}
\newcommand{\bu}{\mathbf{u}}

\newcommand{\largeK}{N}
\newcommand{\sizeDesign}{{N^*}}
\newcommand{\idxDesign}{q}

\newcommand{\btheta}{\boldsymbol{\theta}}

\newcommand{\bgamma}{\boldsymbol{\gamma}}
\newcommand{\bhgamma}{\boldsymbol{\Tilde{\gamma}}}

\newcommand{\argmin}{\arg\!\min} 

\begin{document}

\def\spacingset#1{\renewcommand{\baselinestretch}%
{#1}\small\normalsize} \spacingset{1}

\title{\bf Green LIME: Improving AI Explainability through Design of Experiments}
    \author{Alexandra Stadler\\
    {\footnotesize Institute of Applied Statistics, Johannes Kepler University Linz, Austria}\\
    and \\
    Werner G. M\"uller \\
    {\footnotesize Institute of Applied Statistics, Johannes Kepler University Linz, Austria} \\
    and \\
    Radoslav Harman \\
    {\footnotesize Faculty of Mathematics, Physics and Informatics, Comenius University Bratislava, Slovakia}}
  \maketitle

\begin{abstract}
  
In artificial intelligence (AI), the complexity of many models and processes surpasses human understanding, making it challenging to determine why a specific prediction is made. This lack of transparency is particularly problematic in critical fields like healthcare, where trust in a model's predictions is paramount. As a result, the explainability of machine learning (ML) and other complex models has become a key area of focus.
Efforts to improve model explainability often involve experimenting with AI systems and approximating their behavior through interpretable surrogate mechanisms. However, these procedures can be resource-intensive. Optimal design of experiments, which seeks to maximize the information obtained from a limited number of observations, offers promising methods for improving the efficiency of these explainability techniques.

To demonstrate this potential, we explore Local Interpretable Model-agnostic Explanations (LIME), a widely used method introduced by \cite{ribeiro2016}. LIME provides explanations by generating new data points near the instance of interest and passing them through the model. While effective, this process can be computationally expensive, especially when predictions are costly or require many samples.
LIME is highly versatile and can be applied to a wide range of models and datasets. In this work, we focus on models involving tabular data, regression tasks, and linear models as interpretable local approximations. 

By utilizing optimal design of experiments' techniques, we reduce the number of function evaluations of the complex model, thereby reducing the computational effort of LIME by a significant amount. We consider this modified version of LIME to be energy-efficient or ``green''.
\end{abstract}

\noindent%
{\it Keywords:}  Model Interpretability, Optimal Design of Experiments, Explainable Artificial Intelligence, Post-hoc Explanation, Local Regression.
\vfill

\newpage
\spacingset{1.5}

\section{Introduction}

In artificial intelligence (AI), the complexity of many models and processes exceeds the limits of human understanding, making it difficult to determine the model's reasoning. In certain areas like healthcare, it is not sufficient to provide a model with high accuracy, rather, trust in a model's decisions is crucial \citep[see e.g.][]{london2019}.  The debate on deployment of machine learning models for critical decisions is also subject of legal considerations. The EU’s General Data Protection Regulation (GDPR) includes sections dedicated to the rights of individuals w.r.t. automated decision making. See \cite{kaminski2018} for more information on the scope of the GDPR and its relation to Explainable AI (XAI). Consequently, enhancing the explainability of machine learning (ML) and other complex models has become a vital focus.

Improving model explainability typically involves analyzing AI systems and replicating their behavior using simpler, more understandable mechanisms. However, these approaches can be computationally demanding. As AI continues to play an increasingly significant role in daily life, the judicious use of resources has become a critical topic of discussion. It is essential that, while striving to enhance AI performance, society also prioritizes energy efficiency and computational sustainability. This balance is key to fostering the development of AI technologies that are not only innovative but also ``green'' \citep[see][]{schwartz2020}. The optimal design of experiments, which aims to extract the maximum amount of information from a limited set of observations, presents a promising strategy for increasing the efficiency of these explainability techniques.

This manuscript utilizes two different types of models. One is the complex (ML) model that requires an explanation. The second type is a simple, interpretable model that is used as a surrogate for the complex model to facilitate understanding. The complex model shall subsequently be called the ``primary model'', whereas the simple model used for the explanation will be deemed the ``secondary model''.

In Section \ref{sec:LIME-model-fit}, the secondary model fit is briefly described. Section \ref{sec:LIME-model-fit-sampling} describes the sample generation for the secondary model in detail, including a brief overview of the literature and modifications to the standard LIME algorithm. In the subsequent Section \ref{sec:optimal design}, the topic of D-optimal design for local linear models is introduced. Subsection \ref{subsec:ODE-LIME} describes the application of optimal design approaches to the LIME sampling step in order to make it more efficient. The ideas of this work are illustrated on an example in Section \ref{sec:illustrative-example}. The evaluation of results is discussed in Section \ref{sec:evaluation-metrics} including Subsection \ref{sec:evaluation-results}, which applies the introduced metrics to the aforementioned illustrative example. Lastly, we present a brief discussion and conclusion in Section \ref{sec:discussion}.

\section{Local Linear Models as Explanations (LIME)\label{sec:LIME-model-fit}}

Local Interpretable Model-agnostic Explanation (LIME) is a widely used method in XAI. The method was first introduced by \cite{ribeiro2016} and an open-access implementation in Python is available on GitHub. As the name suggests, LIME is a  method to generate local explanations, i.e. explanations at the instance level, rather than global explanations that cover the entire model domain. The method works on all types of primary models since the technique is based on fitting an interpretable (surrogate) secondary model to new input and output data of the primary model, it is, therefore, model-agnostic w.r.t. the primary model. Interpretability is not well-defined and is subject to the recipient of an explanation. The complexity of the  secondary model in LIME is, thus, modifiable. One way to alter complexity in a secondary model is to limit the number of coefficients. Additionally, it is also possible to use different classes of models (e.g. linear regression, decision tree, etc.) since different types of ``simple'' models might be more understandable to different recipients. 

LIME is applicable to any primary model and relies only on input and output data of the model, hence, it is a post-hoc explanation method. For a taxonomy of explanation methods in different contexts, see e.g. \cite{linardatos2021}.

LIME is highly versatile and can be applied to a wide range of situations and datasets. In this work, we focus on primary models involving tabular data, regression tasks, and linear models as interpretable local approximations (secondary models) to showcase our approach.

We refrain from applying discretization methods to the input variables, as is done in the default method in LIME. We believe that doing so would result in a loss of information in the explanation and that it would not generally make the results more interpretable, even for a general audience.

\begin{figure}
{
    \centering
    \includegraphics[width=0.7\linewidth]{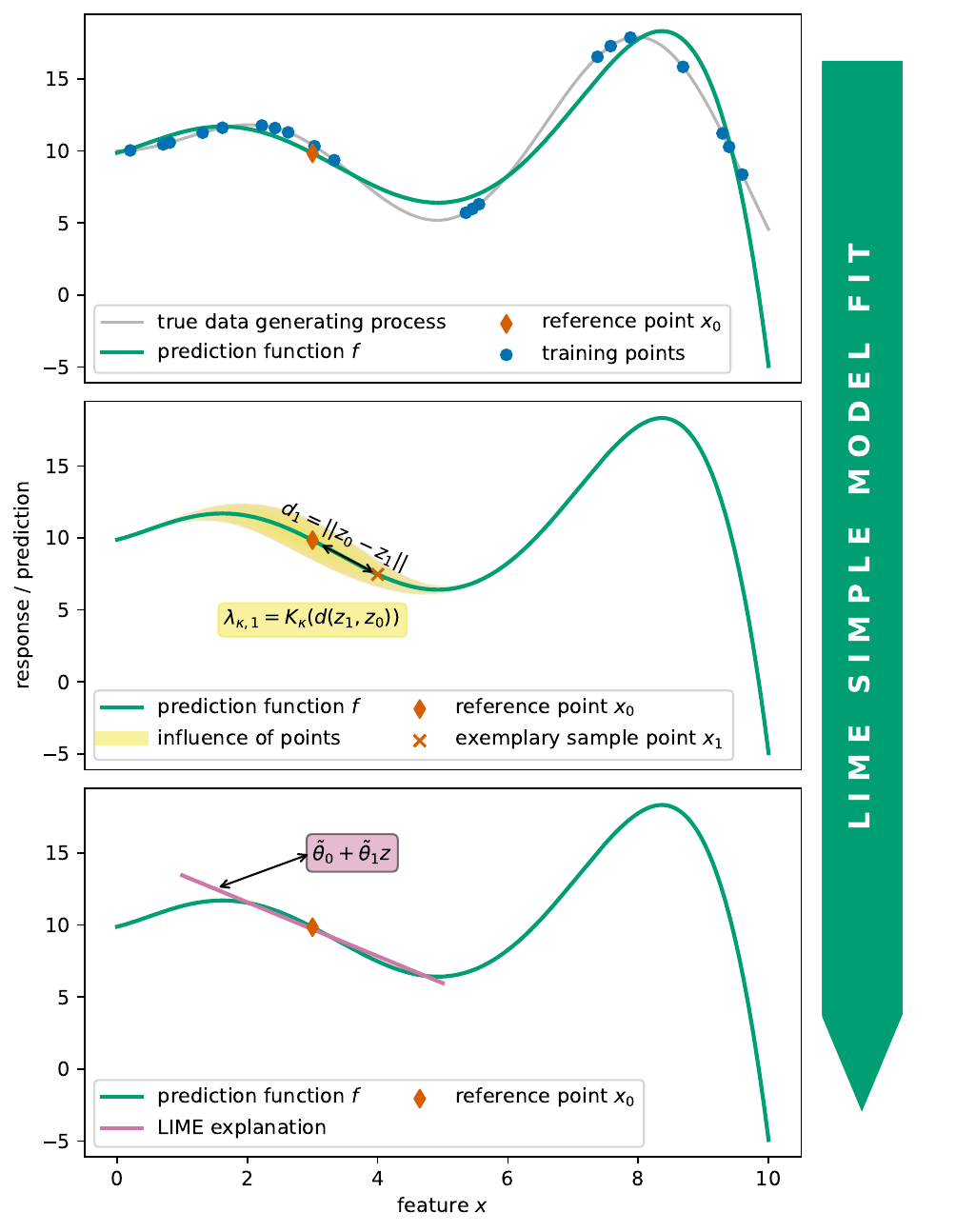}
    \caption{LIME interpretation on an illustrative example}
    \label{fig:toy-example-explanation}
    }
    {\footnotesize
    \setstretch{1.0}
This figure shows the illustrative example presented in \cite{visani2020}. The first plot contains some true data generating process in gray with some training data that was obtained, visualized by the dark blue points. A polynomial primary model (in green) is fit to the training data. This model represents the complex model. Contrary to this example, the functional form and prediction surface is not known in a real application. A reference point at $x_0$ (red diamond) that requires an explanation is selected. In the second plot, the influence of the kernel function and bandwidth is visualized by the yellow shaded area. Points closer to the instance of interest will be weighed more heavily. The distance to an exemplary point at $x_1$ (red cross) is visualized by the black arrow. Lastly, the third plot shows a local linear approximation to the prediction function in the neighborhood of the reference point. The coefficient $\Tilde{\theta}_1$ shows the local influence of feature $x$ on the prediction via its normalization $z = \omega(x)$. Namely, by increasing $x$ close to the instance of interest, the prediction will decrease, i.e. $x$ has a negative impact on the predicted value. Clearly, this is only true in the vicinity of the reference point, i.e. it is true locally. For example, around $x=7,$ an increase in $x$ would increase the predicted value. 
}
\end{figure}

The present section describes the LIME method in a general setting. To facilitate understanding, we present the workflow on an illustrative example in Figure \ref{fig:toy-example-explanation}, originally introduced in \cite{visani2020}. The given dataset is one dimensional and the prediction function is a polynomial of degree 5. Naturally, this primary model does not require an explanation, but it shall be sufficient to visualize the idea of LIME.

Suppose there is a  primary model that produces predictions that shall be explained. Let the primary model's prediction function be denoted by $f$ with an input denoted by some vector $\x \in \mathbb{R}^m$, i.e. with $m$ variables, and an output $y \in \mathbb{R}$. This setup constitutes a regression task. The prediction function $f$ shall be approximated by a linear model in the vicinity of the reference point. The prediction at a point $\x$ is denoted by $y = f(\x).$  The proximity of any point in the input space to the reference point is quantified by a kernel function with a distance metric computed on the instances. The kernel function is dependent on a locality parameter (otherwise called kernel width or bandwidth) that governs the size of the neighborhood.

Let $X \in \mathbb{R}^{n \times m}$ be the training data that was used for fitting the  primary model. Each row in this matrix represents an instance and each column represents a variable, the $i$-th element in the $j$-th column is denoted by $x_{ij}$, with $i = 1, \dots, n$ and $j = 1, \dots, m$. Let $\bar{x}_j$ denote the mean of column $j$ and $s_j$ denote the empirical standard deviation of column $j$, i.e. $s_j = \sqrt{\frac{\sum_{i=1}^n (x_{ij} - \bar{x}_j)^2}{n}}$. We define a standardization function as $\omega_j: \mathbb{R} \to \mathbb{R},~x \mapsto \frac{x - \bar{x}_j}{s_j}.$ In the following, the inverse operation of back transforming to the original input space is also applied and denoted by $\omega_j^{-1}$. Let $\bar{\x}$ denote the vector of means  and $\mathbf{s}$ the vector of standard deviations. The rescaling of an entire instance is similarly denoted by $\omega: \mathbb{R}^m \to \mathbb{R}^m,~\x \mapsto (\x - \bar{\x}) \oslash \mathbf{s},$ where $\oslash$ denotes the elementwise division.

Let $\x_0$ be an instance of interest that may or may not be in the training data. An explanation is computed on the standardized instance $\z_0 = \omega(\x_0)$. The following simple model is used for the approximation:

\begin{equation}
    y = h(\mathbf{z})^\intercal \btheta + \varepsilon,
\label{eqn:approximation-model}
\end{equation}
where $h(\z)^\intercal = \begin{pmatrix}
    1 & \z^\intercal
\end{pmatrix}$, $\btheta \in \mathbb{R}^{m+1}$, and $\varepsilon$ is some remainder that cannot be captured by the approximation. 

The secondary model in Equation \eqref{eqn:approximation-model} shall approximate the function $f$ only in a small neighborhood around $\x_0$. Therefore, a measure of proximity is introduced by a kernel function $K_\kappa$ with hyperparameter $\kappa$. The default kernel function in LIME is given by

\begin{equation}
    K_\kappa: \mathbb{R}^+ \to [0, 1],~d \mapsto \text{exp}\left(- \frac{d^2}{2 \kappa^2 } \right),
    \label{eqn:kernel-function}
\end{equation}
where the input is a distance metric between instances. 
The default distance metric in LIME is the Euclidean distance, which is defined as 
$
    d: \mathbb{R}^m \times \mathbb{R}^m \to \mathbb{R}^+,~(\z_1, \z_2) \mapsto \sqrt{\left(\z_1 - \z_2\right)^\intercal \left(\z_1 - \z_2\right)}.
    \label{eqn:euclidean-distance}
$

Suppose a set of instances $\z_i$, with $i = 1, \dots, \largeK$, is used for the approximation. Then the proximity (referred to as locality) of an instance $\z_i$ to the reference point $\z_0$ is calculated by 

$$
\lambda_{\kappa,i} = K_\kappa\left[ d(\z_i,~\z_0) \right].
$$
Using the set of instances $\z_i$ for the local approximation, the least squares minimization for $\btheta$ is calculated as the solution to 

$$ \Tilde{\btheta} = \argmin_{\btheta} \sum_{i=1}^\largeK \lambda_{\kappa,i} \left( y_i - h(\z_i)^\intercal \btheta \right)^2.  $$
The LIME interpretation for the prediction of $\x_0$ is essentially given by

$$ \begin{pmatrix}
    z_{01} \Tilde{\theta}_1 \\
    z_{02} \Tilde{\theta}_2 \\
    \vdots \\
    z_{0m} \Tilde{\theta}_{m}
\end{pmatrix}.$$
In the accompanying code by \cite{ribeiro2016}, the authors display the effects in order from highest to lowest absolute value. It is clearly visible from this output, that LIME falls into the category of attribution based methods in XAI \citep{agarwal2022a}, i.e. the value of the prediction of $\x_0$ is given as the contribution of the effects from individual variables. 

\section{Neighborhood Sampling in LIME\label{sec:LIME-model-fit-sampling}}

The previous section has discussed secondary model fit in LIME, while omitting a discussion what data $\z_i, i=1, \dots, \largeK$, is used. The generation of this data set is the core topic of this paper. In LIME, the new data is essentially generated by sampling from a multivariate normal distribution, the details of which are discussed in this section.

Let $\mathbf{\Sigma}$ be a diagonal matrix with the squared column-wise standard deviations $s_j,$ $ j= 1, \dots, m$, of the primary model's training data as entries. This will be the covariance matrix of the training data for the secondary model. Notice that this matrix implies independence between the columns, i.e. variables/features, in the secondary model's training data.

A sample of size $\largeK \in \mathbb{N}\backslash\{0\}$ shall be sampled for the model fit in the local interpretable secondary model. The new data is randomly sampled as $\breve{\x}_i \sim \mathcal{N}(\x_0, \mathbf{\Sigma}),$ for $i=2, \dots, \largeK$, and $\breve{\x}_1 = \x_0$. The new data points have the same dimension as the instance of interest $\x_0$. This operation is called ``sampling around the instance of interest''. Alternatively, scaling according to the training data's mean and standard deviation could be applied. This would correspond exactly to sampling from a normal distribution centered at the training data's vector of mean values. There are arguments for both approaches, but we prefer to work with sampling around the instance of interest for easier comparisons with our method. We feel that this is more useful, since the kernel will lead to very small weight for instances that are far away from the reference point.

To obtain responses (sometimes called labels) for the simple secondary model fit, the prediction function $f$ is applied to the new data, i.e. $y_i = f(\breve{\x}_i).$ As described in Section \ref{sec:LIME-model-fit}, the secondary model is fit to rescaled data, hence the new sample is rescaled by $\z_i = \omega(\breve{\x}_i).$ The locality weights are computed by first calculating the distances $d_i = d(\z_i, \z_0)$, and then applying the kernel function so that $\lambda_{\kappa,i} = K_\kappa(d_i).$ Notice that $d_1$ is always 0 and hence $\lambda_{\kappa,1} = 1$, since the first point in the new sample is equal to the reference point.

\begin{figure}
{
    \centering
    \includegraphics[width=0.8\linewidth]{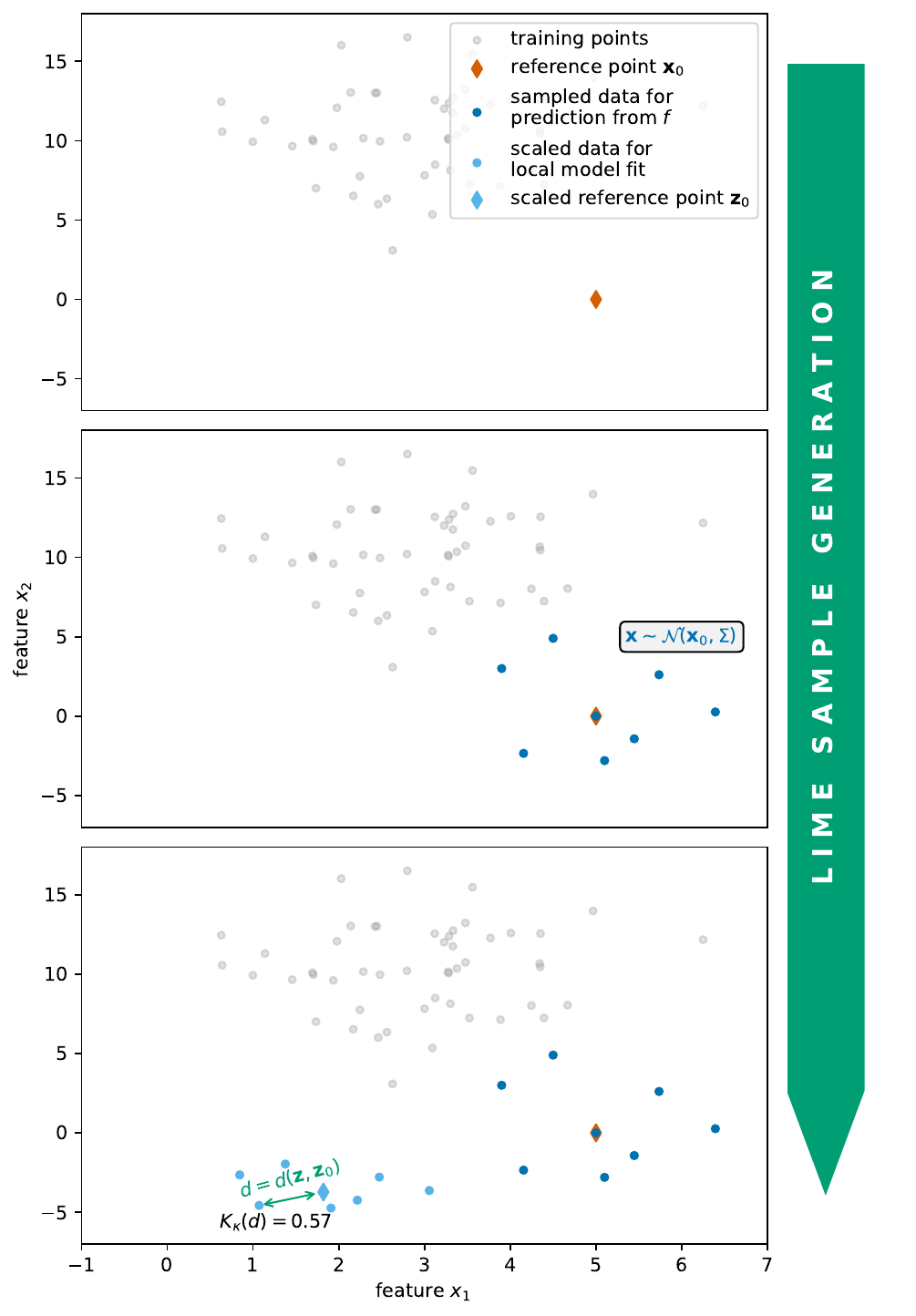}
    \caption{LIME sample generation for two features}
    \label{fig:lime-sampling}
    }
{\footnotesize
This figure is a representation of the new sample generation in LIME. In the first plot, training data for the primary model is depicted. The data set contains two features. In the second plot, new data (in dark blue) for the simple secondary model is randomly sampled from a normal distribution centered at the reference point $\x_0$ and with standard errors calculated on the training data. This data is used for obtaining the responses $y = f(\x)$. The third plot shows the rescaled data points and reference point in light blue. The distance between the rescaled reference point and a new rescaled sample point is visualized by a green arrow and the corresponding weight is calculated by the kernel function $K_\kappa$ with $\kappa = \frac{3}{4} \sqrt{2}$.
}
\end{figure}

A visualization of an example for sampling in the LIME workflow is given in Figure \ref{fig:lime-sampling}.

\subsection{Modifications to LIME\label{subsec:modifications}}

The scientific community has shown quite a lot of interest in LIME, with over 1600 articles containing ``Local interpretable model-agnostic explanations'' in the article title, keywords or abstract as found by a simple search on Scopus (June 6\textsuperscript{th} 2025). The literature can be divided into applications and modifications of LIME. In particular, the overall sampling step, the choice of kernel width, and the size of the sample are subject to ample discussion. Some approaches include a subselection of randomly sampled data in classification tasks \citep{Saadatfar2024}, agglomerative hierarchical clustering of the original training data and the K-nearest-neighbor algorithm \citep{zafar2021}, training an autoencoder on the training data and using embeddings to measure the distance between instances \citep{Shankaranarayana2019}, adaptively determining the sample size such that the results of LIME are stable \citep{zhou2021}, optimizing the kernel width w.r.t. a trade-off between criteria \citep{visani2020}.

The most frequent point of critique is the instability of LIME \citep{visani2022, visani2020, zafar2021, Shankaranarayana2019, Saadatfar2024, jiang2022}. Instability in LIME refers to the fact that the method leads to different results for multiple runs of the algorithm. This stems from the fact that sampling of neighborhood data is completely random, hence the training data for the interpretable secondary model is different at each run (except if a random seed is set in advance). Depending on the sample size that is specified and the size of the neighborhood, i.e. the bandwidth, the method can even yield conflicting results between runs. A variable could for example positively affect the prediction in one run and have a negative impact in another run. This type of variability in the results is problematic, since one of the main proposed purposes of LIME is to increase trust in a model \citep{ribeiro2016}. 

Another fundamental question that arises for researchers and practitioners alike is the specification of the kernel width $\kappa$, \cite[see e.g.][]{molnar2022}. In a scenario where the prediction surface of a complex primary model is not clear, it is difficult, if not impossible, to specify a reasonable kernel width. The neighborhood must be specified such that a linear secondary model can approximate the primary model reasonably well. Hence, for a highly non-linear prediction surface, the kernel width should be small. Additionally, the degree of nonlinearity is not necessarily uniform across the whole input space, i.e. knowledge about the prediction surface in one region does not translate to other regions. \cite{ribeiro2016} specify a default parameter of $\kappa = \frac{3}{4} \sqrt{m}$ in their code, we believe that this value has proven most useful in the developers' experiments. The specification of a reasonable kernel width is still subject of discussion among the scientific community, see e.g. \cite{visani2020}, who optimize the kernel width for a trade-off of two criteria. Such an approach encompasses multiple runs of LIME in order to compute statistical quantities on the obtained results, and is, therefore, costly. 

Lastly, the sampling approach by \cite{ribeiro2016} relies on the prediction of many instances, and thus, execution of the prediction function $f$. Many instances that are sampled are ultimately irrelevant to the local secondary model, since the distance to the instance of interest is too large, resulting in a locality weight that is barely larger than zero. This means that some (or even many) instances that are sampled and evaluated give no information about the model's behavior in the neighborhood of the reference point. This may seem like a minor issue, since the sample size can be arbitrarily increased, however, the resulting energy consumption and prolonged computational runtime can be meaningful if an algorithm as popular as LIME is widely applied. \cite{desislavov2023} argue that the training of a model may seem like the costly part compared to deployment. However, in most cases a model is trained only once but the cost of inference is subject to a multiplicative factor, which can make it significantly more costly during deployment. This is not surprising, but emphasizes that multiple evaluations of $f$ scale-up negatively. It may even be the case that for some applications an explanation will be (or has to be) provided immediately upon prediction of an instance. This would result in a considerable increase of evaluations of $f$ (e.g. by a factor or 5000 if this is the sample required).

The situation becomes even more dire when the primary model to be explained has a prediction function that is costly to evaluate even in one run. This is not only an annoyance for practitioners, but consumes even more energy when additional units are sampled that are ultimately not required. The current trend in research is to make AI ``greener'', i.e. to decrease energy consumption and carbon footprint of AI \citep{schwartz2020}. The goal of this paper is to showcase the use of optimal design of experiments in the context of LIME in order to decrease the number of function evaluations needed, while maintaining comparable results. The application at hand is a starting point for the use of optimal design in efficient data collection in XAI.

\section{Optimal Design of Experiments for Local Linear Models\label{sec:optimal design}}

Optimal experimental design is a subfield of statistics that aims to improve data collection given restrictions from the real world, be it budgetary, time or other limitations that need to be considered. While initially, the field has worked on classical problems in laboratory and industrial settings, the advances in modern data analysis and the increased interest and facility of data collection have pushed the field towards expanding the methodology to evermore challenging tasks. The word ``experiment'' should not only be understood in the ``traditional'' sense, but should rather denote any task of data collection. For an overview of modern optimal experimental design and its roots see \cite{lopez-fidalgo2023} and for a Bayesian viewpoint see \cite{huan2024}.

Using linear models as local approximations to complicated functions is not a new idea, and neither is the design of experiments for such local models, although the method might appear under different names, e.g. moving local regression, local fitting, linear surrogate model, proxy model, etc.. We will mainly refer to the article by \cite{fedorov1999}, which provides the foundation of design of experiments in local linear models building up upon \cite{muller1991, muller1996}. 
This includes the well-known D- and A-optimal design criteria for this context, as well as the corresponding sensitivity functions (directional derivatives) that can be helpful for swift optimization of the criterion value. A newer publication by \cite{fisher2013} follows similar ideas and extends the literature to cases where a design is sought that maximizes the D-criterion for a subset of parameters in the model.

The present section introduces fundamentals on the design of experiments for local linear regression models, while maintaining similarity to the notation in Section \ref{sec:LIME-model-fit}. In particular, $f$ shall be a function that will be approximated in the neighborhood of a point $\z_0 = \omega(\x_0)$.

Suppose the following model holds

\begin{equation}
y_i = \eta(\z_i) + \varepsilon_i,    
\label{eqn:local-model}
\end{equation}
where $\eta(\z)$ is a smooth response function and $\varepsilon_i$ are homoskedastic independent errors with zero mean for $i = 1, \dots, \largeK$. The approximation of the smooth response function is performed via linear regression as in

\begin{equation*}
\eta(\z_i) = h(\bu_i)^\intercal \bgamma + g(\bu_i)^\intercal \boldsymbol{\zeta},    
\end{equation*}
where $\bu_i = \z_i - \z_0$, $\bgamma \in \mathbb{R}^{(m+1)}$ and $\boldsymbol{\zeta} \in \mathbb{R}^{s}, s \in \mathbb{N},$ which is similar to the model in \cite{fedorov1999}. The term $g(\bu_i)^\intercal \boldsymbol{\zeta}$ is the so-called remainder term in this model. It quantifies the error induced by the approximation. In the following, we assume that the bias is negligible and omit the term from the model. 

Assume that the following linear model is used for estimation 

\begin{equation*}
    y_i = h(\bu_i)^\intercal \bgamma + \varepsilon_i,
\label{eqn:model-single-point}
\end{equation*}
where $\bu_i = \z_i - \z_0 $  and in our case $h(\bu_i)^\intercal = \begin{pmatrix}
    1 & \bu_i^\intercal
\end{pmatrix},$ for $i = 1, \dots, \largeK.$
The vector of coefficients $\boldsymbol{\Tilde{\gamma}}$ is then obtained via 

\begin{equation*}
    \bhgamma = \argmin_{\bgamma} \sum_{i=1}^\largeK \lambda_{\kappa,i} \left( y_i - h(\bu_i)^\intercal \bgamma \right)^2,
    \label{eqn:estimation_theta}
\end{equation*}
where $\lambda_{\kappa,i}$ are weights defined according to some kernel function and distance metric. For this particular application, we aim at computing the weights in the same manner as in LIME. Hence, the weights are given by $\lambda_{\kappa,i} = K_\kappa(d(\z_i, \z_0)),$ for some kernel width $\kappa$.

Let an approximate design be defined by $\xi = \{\z_\idxDesign, p_\idxDesign\}_{q=1}^\sizeDesign$ with $0 \leq p_\idxDesign \leq 1$ and $\sum_{\idxDesign=1}^\sizeDesign p_\idxDesign = 1$, where $N^* \in \mathbb{N}\backslash\{0\}$ denotes the size of the support of design $\xi$.
Define the following matrices

\begin{equation*}
    \mathbf{M}_{11}(\xi) = \sum_{\idxDesign=1}^\sizeDesign p_\idxDesign \lambda_{\kappa,\idxDesign} h(\bu_\idxDesign) h(\bu_\idxDesign)^\intercal, 
    \label{eqn:M11}
\end{equation*}

\begin{equation*}
    \widetilde{\mathbf{M}}(\xi) = \sum_{\idxDesign=1}^\sizeDesign p_\idxDesign \lambda_{\kappa,\idxDesign}^2  h(\bu_\idxDesign) h(\bu_\idxDesign)^\intercal. 
    \label{eqn:M_tilda}
\end{equation*}
Let the normalized mean squared error matrix (assuming that the bias is zero or at least negligible) for $\bhgamma$ be 

\begin{equation*}
\R(\xi) = \mathbf{M}_{11}^{-1}(\xi) \widetilde{\mathbf{M}}(\xi) \mathbf{M}_{11}^{-1}(\xi),
\label{eqn:normalized-MSE-matrix}
\end{equation*}
by extension of the results of \cite{fedorov1999}.

We aim to optimize the D-criterion, which is given by 

\begin{equation*}
    \Psi_D[\R(\xi)] = \text{log} | \R(\xi) |. 
    \label{eqn:D-criterion}
\end{equation*}
Let the set of all admissible approximate designs be $\Xi$, then a D-optimal design is defined as 

$$
\xi^* = \argmin_{\xi \in \Xi} \Psi[\R(\xi)].
$$
We assume that $\mathcal{Z}$ is the compact set of all permissible points in the experiment and $K_\kappa$, $h$ are continuous functions. \cite{fedorov1999} state some further necessary conditions for their theory.

\subsection{Optimal Design for the LIME Sampling Step\label{subsec:ODE-LIME}}

\begin{figure}[htb]
    \centering
    \includegraphics[width=0.5\linewidth]{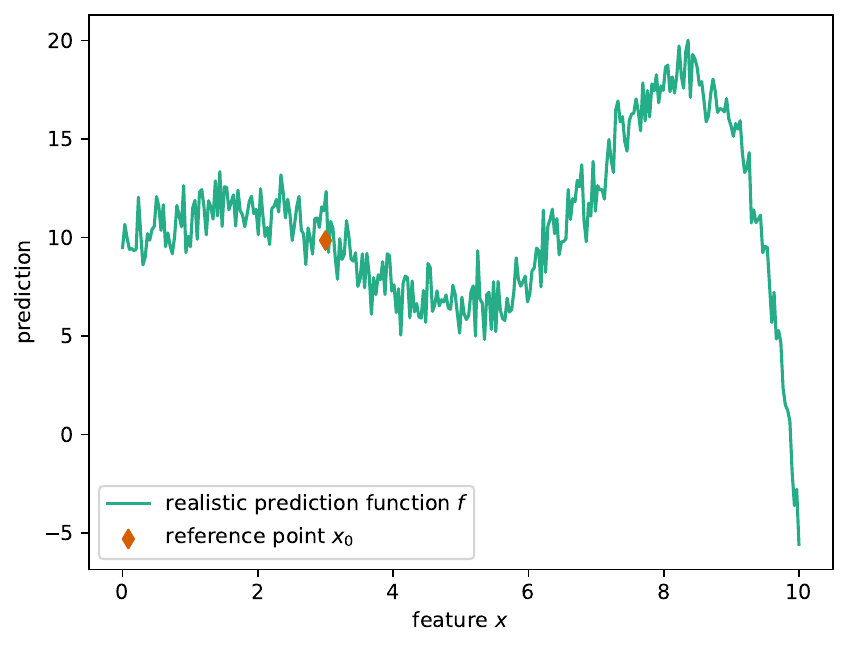}
    \caption{Realistic prediction function for the illustrative example}
    \label{fig:toy-example-realistic}
\end{figure}

The function $f$ to be approximated, i.e., the primary model's prediction function, is typically not subject to randomness. Hence, Equation \eqref{eqn:local-model}  does not require an additive random term, as the only error that arises from an approximation at $\z_0 = \omega(\x_0)$ is bias due to model misspecification. However, in complex models, we expect the prediction surface to be (highly) ragged, i.e., for small changes in the input, the response will be different. We visualize this expected behavior of the prediction function in the example in Figure \ref{fig:toy-example-realistic}.

Thus, it is not useful to compute a weighted average over repeated evaluations of the deterministic response at one design point. This is also the reason why we omitted a subindex in the local linear model as opposed to the formulas in \cite{fedorov1999}.

The task of providing an explanation remains the same: we would like to capture the average behavior in the neighborhood of the reference point. This is also in contrast to computing the gradient at position $\x_0$. The gradient is the most accurate approximation at point $\x_0$ in an infinitesimal neighborhood, however, it may not be an appropriate representation, of how variables impact the prediction in a typical neighborhood of $\x_0$. There is quite obviously a trade-off between capturing the influence of variables at $\x_0$ and in the neighborhood of this point. 

Another difference in the model above is that the approximation is computed on the variables $\bu_i$, i.e. the difference between the reference point and other points. However, this is not an issue, since the input variables can be arbitrarily rescaled, so long as the responses $y_i$ are computed on the original feature space.

In the LIME setting, the center point is always included in the new sample for the secondary model fit. This somewhat counterbalances misspecification should the neighborhood be nonlinear and provides some robustness for the method.

However, for optimal design considerations, there is an additional reason for including the center point in the new sample. In a one-dimensional example without bias term and including a center point \cite{fedorov1999} have shown that the optimal design will be symmetric around the reference point and will put equal weight at the boundaries of the feature values, i.e. it will move the optimal design points as far away from the reference point as possible. As this is impractical, since these design points will have almost zero locality weight and, therefore, little information, \cite{fedorov1999} recommend allocating a regularizing weight to the reference point. While this strictly sacrifices D-optimality it will lead to more stable designs.

Thus, in accordance with the original LIME setting, we will follow this suggestion by fixing some minor weight $\delta$ at the reference point. Then the only factors determining the optimal design are the kernel width $\kappa$ and the dimension $m$ of features. The location of the reference point is arbitrary and our design variables $\bu_i = \z_i - \z_0$ are such that the reference point corresponds to the zero vector, which shall be called ``center point''. Suppose $N+1$ units shall be sampled in total, then a design weight of $\delta = \frac{1}{N+1}$ will be allocated to the center point. This is essentially a restriction on the admissible set of designs $\Xi$.

As mentioned, in one dimension, the D-optimal design with regularizing weight $\delta$ will be a three point design symmetric around the center point, i.e. it will be

\begin{equation*}
\xi^* = \begin{Bmatrix}
    -u & 0 & u \\
    \frac{1-\delta}{2} & \delta & \frac{1-\delta}{2}
\end{Bmatrix}.
\end{equation*}

As by the results in \cite{fedorov1999}, the D-criterion can conveniently be expressed as a function of the kernel width and the scalar design point $u$, which can be interpreted as the optimal distance to the reference point. The function can be written as

\begin{equation}
    D(u, \kappa) =  \frac{(1-\delta)\lambda_\kappa(u)^2 + \delta}{((1-\delta)  \lambda_\kappa(u) + \delta)^2 (1- \delta) u^2}. 
\label{eqn:D-criterion-one-dimensional}
\end{equation}
This function can be optimized in the scalar value $u$ without much computational effort, save for the definition of reasonable optimization bounds such that it is well-behaved. Even this can be further simplified by leveraging the properties of the RBF-Kernel and Euclidean distance. In this setting, the locality weight is computed as

\begin{equation*}
    \lambda_\kappa(u) = \exp\left(-\frac{u^2}{2 \kappa^2}\right) = \exp\left(-\frac12 \left(\frac{u}{\kappa}\right)^2\right) = \lambda_1\left(\frac{u}{\kappa}\right). 
\end{equation*}
It follows that

\begin{equation*}
    D(u, \kappa) =  \frac{(1-\delta)\lambda_1\left(\frac{u}{\kappa}\right)^2 + \delta}{((1-\delta)  \lambda_1\left(\frac{u}{\kappa}\right) + \delta)^2 (1- \delta) \left(\frac{u}{\kappa}\right)^2 \kappa^2} \propto D\left(\frac{u}{\kappa}, 1\right). 
\end{equation*}
It is thus sufficient to obtain $u^* = \argmin_u D(u, 1)$ and rescale to $\kappa u^*$ to obtain the optimal value for a kernel width $\kappa$. 

Note that we assume $f$ induces equal costs for the prediction of different units. Hence, our resources are solely restricted by the size of the sample, i.e. the number of experimental units available, and no further resource constraints need to be considered.

It might easily be the case that the requested sample size is larger than the support of the approximate design. This can oftentimes lead to replicates at design points. Since there are no true replicates in this setting (due to the deterministic nature of $f$), we propose a slight variation of the procedure. 
Instead of including a design point multiple times, the sample is ``enhanced'' by allocating a subset of the available units to each design point, proportional to the design weight, and generating a new sample point by adding some small Gaussian noise to the design point. We call this approach ``jittering'' (around the design point), which enables specification of a design with arbitrary size (larger than the support of the design). Others have proposed similar ideas to increase robustness against model misspecification, see e.g. \cite{waite2022, wiens2024}.

Suppose $\largeK +1$ samples should be included in the final sample and $\sizeDesign \leq \largeK +1$. Then one unit is assigned to the center point and $\largeK$ units are distributed to the remaining support points w.r.t. to their design weight via the efficient rounding technique discussed in \cite{pukelsheim1992}. Suppose this results in sample allocations of $\largeK_\idxDesign$. Then all support points $\mathbf{v}_\idxDesign$ are included in the design and the remaining sample points are generated by adding a small Gaussian noise with zero mean and a small variance to the support points. Denote this sample by $\z_i$, $i = 2, \dots, \largeK$ and add $\z_1 = \z_0$ as is done in the LIME methodology.

\begin{figure}[htb]
{
    \centering
    \includegraphics[width=\linewidth]{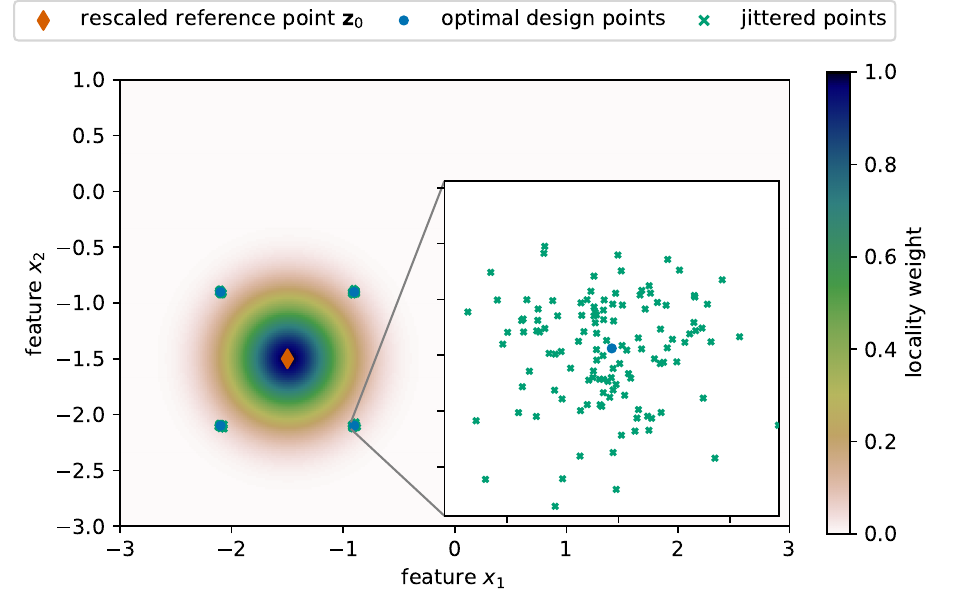}
    \caption{Enhanced sample by jittering around the support points}
    \label{fig:jittering}
}
{\footnotesize
    This figure illustrates a sample generated by our method. A reference point is marked by a red diamond. The neighborhood is characterized by the Euclidean distance and RBF kernel, as given in Equation \eqref{eqn:kernel-function}. The corresponding hyperparameter $\kappa$ was set to $\frac{1}{4} \sqrt{2}$. A (near-) optimal design with center point weight is illustrated by blue points surrounding the reference point. The support points are positioned such that the locality weight is at approximately $\frac{1}{18}$. The total number of samples is set to 501, of which one is assigned to the rescaled reference point and four are apportioned to the remaining support points. The remaining 496  points are allocated to the support points' respective samples, which are proportional to the design weight. These points are then perturbed by Gaussian random noise with mean 0 and variance $0.01^2$.
}
\end{figure}

For an illustration of our method with two input features, see Figure \ref{fig:jittering}.

\section{Explanations on the Illustrative Example\label{sec:illustrative-example}}

Let us now proceed again with the example from \cite{visani2020}. This example is suitable for visualization since it contains only one feature and thus the prediction function is a curve, hence, we may easily be able to derive some intuition from graphs. 

\begin{figure}[htb]
    \centering
    \includegraphics[width=0.8\linewidth]{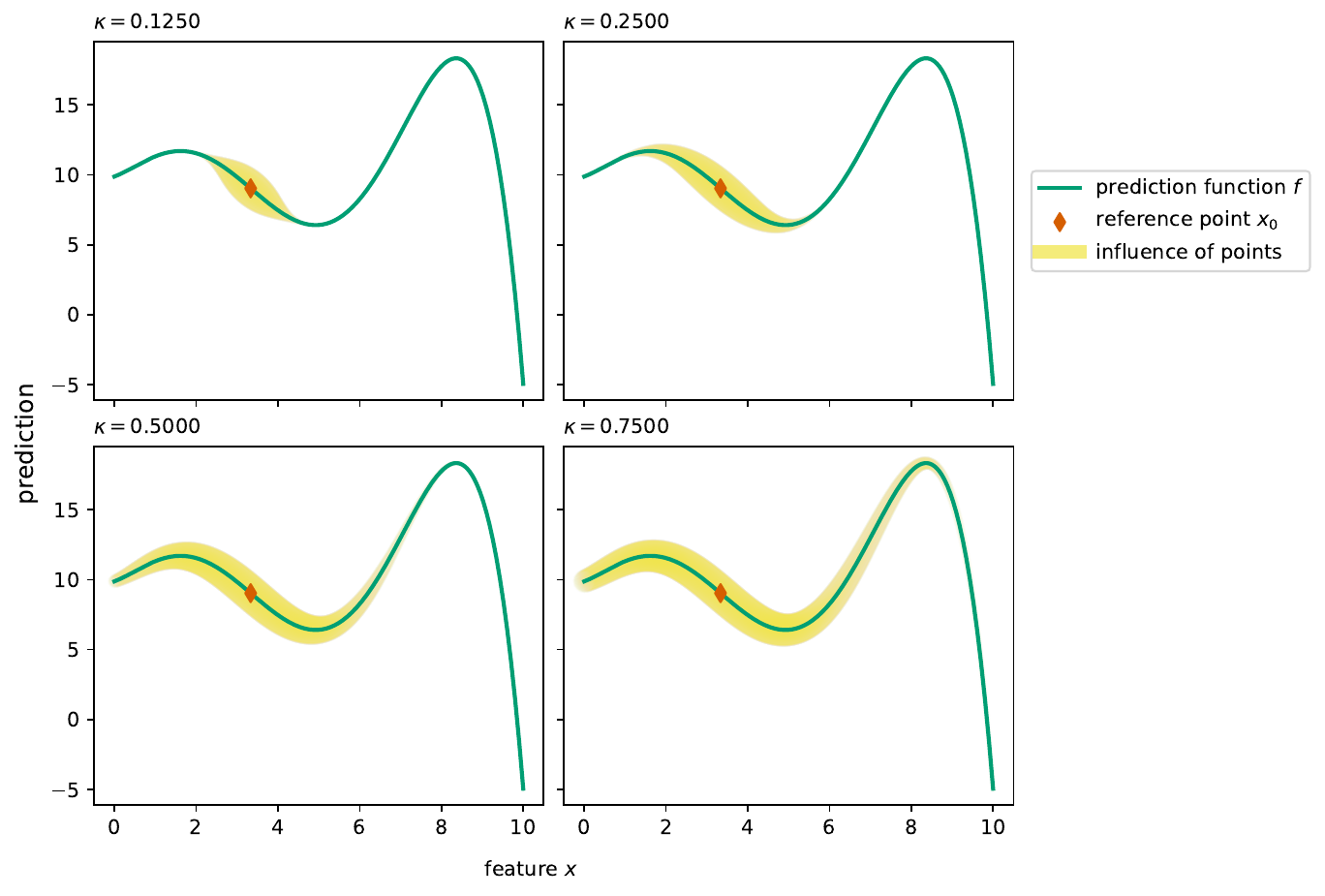}
    \caption{Influence of different kernel widths}
    \label{fig:kernel-widths}
\end{figure}

At first, we would like to point out some particularities in this example. For one, we visualize the size of the neighborhood of a reference point (training point 10 in this case) w.r.t. the chosen kernel width $\kappa$ in Figure \ref{fig:kernel-widths}. It is immediately clear from this figure that the default kernel width of $\kappa = \frac{3}{4} \sqrt{1}$ is not suitable for this example, perhaps not for any example in one dimension. Even a kernel width of $\frac12$ seems to be too large, since the prediction function in this neighborhood is still quite non-linear. Some value close to $\frac14$ seems reasonable for this reference point. \cite{visani2020} suggest $\kappa = 0.296$ as a reasonable kernel according to the specifications of their method. 

The choice of kernel width is not subject of this article, but naturally, the quality of the results heavily depends on its reasonable specification. We, thus, operate under the assumption of well-specified kernel widths. The choice will be discussed in the following. 

First, we specify some reference points that can be utilized to generate exemplary explanations. The choice of such points is completely arbitrary. It could be of interest to explain predictions from the training set, or perhaps from an existing test set. It could also be the case that predictions on the whole region of input features shall be systematically explained and some theoretical instances are produced by users.

\begin{figure}[htb]
    \centering
    \includegraphics[width=0.6\linewidth]{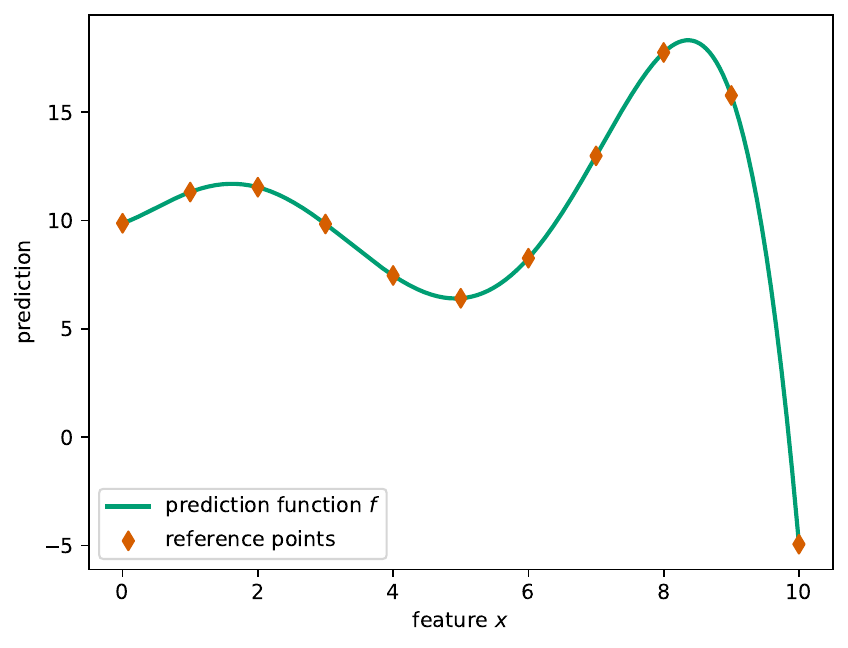}
    \caption{Reference points in the illustrative example}
    \label{fig:grid-11}
\end{figure}

Since the choice is arbitrary, we opt to place reference points on a regular grid on the interval $[0, 10]$. We thus mark 11 equally spaced points in this interval as the reference points, the predictions of which shall be explained. The points are visualized in Figure \ref{fig:grid-11}.

We aim to apply the optimal design approach to this example, while staying otherwise consistent to the LIME approach. However, we do change the default simple secondary model in LIME, which is a ridge regression model with penalty 1. Since we have no desire to shrink the coefficient for the slope in this example, we use a local linear model without penalty. 

Otherwise, we apply the following procedure to generate a sample for the secondary model fit. At first a reference point $\x_0$ out of the 11 points is rescaled to $\z_0 = \omega(\x_0)$.

Then, a small design weight $\delta = \frac{1}{N + 1}$ is allocated to the reference point. The scalar optimal distance from the rescaled reference point for a kernel width of $1$ is computed as by Equation \eqref{eqn:D-criterion-one-dimensional}. This is a bounded optimization task in a scalar function, which is fairly easy to solve, hence, the solution is almost instantaneously obtained. The optimal distance is then rescaled linearly by multiplying with the kernel width $\kappa$. The design points are then placed symmetrically around the rescaled reference point. The remaining design weight $\frac{N}{N+1}$ is divided equally among the two points. This constitutes the 
design, which will be used as a basis to generate the final sample.

The three support points in the design are included into the new sample for the secondary model fit. The remaining $N-2$ design units are added to the sample by jittering around the support (excluding the center point) relative to the design weights with a variance of $0.01^2$. 

As stated before, one critical point in the application of LIME is the specification of an appropriate kernel width (per reference point). In order to generate reasonable results, we rely on what we will subsequently call the ``optimal kernel widths''. We compute these kernel widths according to the approach by \cite{visani2020}. The optimization of the kernel width proceeds by repeated execution of LIME under different kernel width settings, which is inefficient. Assumption of a costly prediction function $f$, would naturally render this approach inappropriate.

The evaluation of results in this application is non-trivial. Hence, we first introduce it in the subsequent section and proceed with some metrics and a discussion afterward.

\section{Evaluation Metrics\label{sec:evaluation-metrics}}

Assessing the quality of explanations in the domain of XAI is a challenging task. Various criteria of quality have been proposed with many metrics for each aspect making survey articles like \cite{barredo-arrieta2020} and  \cite{pawlicki2024} a necessity to develop an understanding of the issue. 

It is not clear what requirements an explanation must fulfill in order to be judged good, suitable or even just useful and \cite{barredo-arrieta2020} highlight that the field has not reached a general agreement on what an explanation is, but emphasize that the answer must be dependent on the audience.

This has also enticed researchers to develop software that encapsulates a number of evaluation metrics, see e.g. Quantus \citep{hedström2023} or OpenXAI \citep{agarwal2022b}. According to \cite{pawlicki2024} the criterion that is mentioned most frequently is fidelity. \cite{ribeiro2016} discuss local fidelity stating that an explanation must be in line with the complex primary model's behavior in the neighborhood of the reference point. The authors also emphasize that there is a trade-off between fidelity and interpretability, since the complexity (e.g. number of coefficients) of the explanation must be significantly limited to serve its purpose. This is, of course, only true if the primary process is too complex for a simple secondary model to be captured (see also \citeauthor{barredo-arrieta2020}~\citeyear{barredo-arrieta2020}, and sources therein).

Another property that is frequently discussed in the context of LIME is stability. We have already discussed issues with this property in Subsection \ref{subsec:modifications}.  Note, that stability of LIME is especially questionable in settings where only a small number of units can be sampled, as by our assumption. 

Since the assessment of the quality of an explanation is not straight-forward, it is in turn not clear how to compare different explanation techniques to one another. Our approach largely relies on the overall method of LIME while only changing the sample generation step. Thus, some properties of LIME are preserved in any case, e.g. interpretability. 

Some common evaluation criteria rely on comparing which (important) features are included in the resulting explanations. One could for instance evaluate the quality of different explanation methods by simulating an example where only a subset of the features have an influence on the output of the primary model and compare the frequency with which they are included (or left out) in the explanation. This is not a suitable approach for this example, since there is only one feature and no feature selection occurs at all. 

Typically, results from a regression model are evaluated by the mean squared error (over some test set). This is in line with the estimation, which proceeds by minimizing the squared deviations from the response in classical linear regression. In the local linear model, a similar loss is minimized, the difference lies in importance of locality w.r.t. the reference point. We thus propose a metric that follows a similar principle and call it the normalized weighted integrated squared error (NWISE). 

Let the prediction function of the simple secondary model be denoted by $g: \mathbb{R}^{m+1} \to \mathbb{R}.$ Given some test set $\mathcal{X}_T$, we define the NWISE as

\begin{equation*}
    \text{NWISE}(g, f| \x_0) = \frac{\int_{\mathcal{X}_T} K_\kappa \left\{d\left[\omega(\x), \omega(\x_0)\right] \right\} \left[ g(\omega(\x)) - f(\x) \right]^2 \text{d}\x}{\int_{\mathcal{X}_T} K_\kappa \left\{d\left[\omega(\x), \omega(\x_0)\right] \right\}  \text{d}\x},
    \label{eqn:NWISE}
\end{equation*}
with bandwidth $\kappa$, distance function $d$ and kernel function $K_\kappa.$
Naturally, the kernel function and kernel width should be chosen as explained above.

One sign of quality of an explanation is the similarity of predicted values between the complex primary model and the simple secondary model. If the  secondary model is a good surrogate for the primary model, the secondary model should frequently predict similar values in the vicinity of the reference point. We attempt to capture this property by a weighted correlation between the predictions of the two models. 

Suppose the predicted values of functions $g$ and $f$ are denoted by $\hat{y}_g(\x)$ and $\hat{y}_f(\x)$ and the covariance between the predicted values is computed w.r.t. the proximity to the reference point. Then we propose the following notion of weighted covariance over some test set $\mathcal{X}_T$

\begin{equation*}
    \text{COV}(g, f|x_0) =  \frac{\int_{\x \in \mathcal{X}_T} K_\kappa \left\{d\left[\omega(\x), \omega(\x_0)\right]\right\}  \left[\hat{y}_g(\x) -  \overline{y}_g(\x) \right]\left[ \hat{y}_f(\x) -  \overline{y}_f(\x) \right] \text{d}\x }{\int_{\x \in \mathcal{X}_T} K_\kappa \left\{d\left[\omega(\x), \omega(\x_0)\right]\right\} \text{d}\x},
\end{equation*}
where $ \overline{y}_g(\x) = \frac{\int_{\x \in \mathcal{X}_T}  K_\kappa \left\{d\left[\omega(\x), \omega(\x_0)\right]\right\} \hat{y}_g(\x) \text{d}\x}{\int_{\x \in \mathcal{X}_T   } K_\kappa \left\{d\left[\omega(\x), \omega(\x_0)\right]\right\} \text{d}\x} $ and similarly for $\overline{y}_f(\x)$
and calculate the correlation $\text{CORR}(g,f | \x_0)$ accordingly.

In this particular case with the simple secondary model $g$, we have $\hat{y}_g(\x) = g(\omega(\x))$ and $\hat{y}_f(\x) = f(\x).$

\subsection{The Illustrative Example Revisited\label{sec:evaluation-results}}

We compare results on explanations over the illustrative example for the original LIME method and our approach (denoted by ODE). Each reference point in Figure \ref{fig:grid-11} is used to generate an explanation in 100 distinct runs. The respective kernel widths are chosen according to the method by \cite{visani2020} and the optimal design is enriched by the jittering approach described in Section \ref{subsec:ODE-LIME}. We generate samples of size 11 for the LIME and ODE approach.

We approximate the NWISE and CORR by summing over a discretized grid of 1001 equally spaced values over $[0, 10]$ instead of computing the integral in each run. 

In this illustrative example, the LIME approach results in worse or equally good outcomes on average for almost all units, see Table \ref{tab:results-optimal-kernels-100-repetitions} for the average (approximated) NWISE and CORR over the 100 runs. 

\begin{table}[h]
    \centering
    \begin{tabular}{crc|rr|rr}
            \hline
			\hline
            & & & \multicolumn{2}{c}{NWISE} & \multicolumn{2}{|c}{CORR} \\
		\hline
			& unit & $\kappa$ & LIME & ODE & LIME & ODE \\ 
			\hline
                & 0 & 0.0601 & 16,429 & \textbf{16,252} & 0.8586 & \textbf{0.9983} \\
			& 1 & 0.1079 & 16,464 & \textbf{16,244} & 0.9154 & \textbf{0.9535} \\
			& 2 & 0.0585 & 16,720 & \textbf{16,553} & 0.8918 & \textbf{0.9488} \\
			& 3 & 0.2854 & \textbf{19,137} & 20,546 & \textbf{0.9478} & \textbf{0.9478} \\
			& 4 & 0.1780 & 26,692 & \textbf{23,679} & \textbf{0.9492} & \textbf{0.9492} \\
			& 5 & 0.0100 & 34,171 & \textbf{34,149} & 0.3547 & \textbf{0.9587} \\
			& 6 & 0.1807 & 24,572 & \textbf{20,634} & \textbf{0.9485} & \textbf{0.9485} \\
			& 7 & 0.2490 & \textbf{30,418} & 30,991 & \textbf{0.9495} & \textbf{0.9495} \\
			& 8 & 0.0612 & 66,340 & \textbf{61,729} & 0.7795 & \textbf{0.9506} \\
			& 9 & 0.0731 & 48,558 & \textbf{34,245} & 0.9312 & \textbf{0.9502} \\
			& 10 & 0.1504 & \textbf{174,770} & 316,728 & \textbf{0.9722} & \textbf{0.9722} \\
            \hline
			\hline
    \end{tabular}
    \caption{Results for the illustrative example (averaged over 100 runs)}
    \label{tab:results-optimal-kernels-100-repetitions}
\end{table}

Note that there is an unexpected deviation for the last unit due to a boundary effect occurring at this position.

The given example has a very smooth surface, hence the approximation cannot benefit as much from the jittering approach as in a typical example. We believe that in a true application, the differences between LIME and ODE are more pronounced.

\begin{figure}[h]
    \centering
    \includegraphics[width=0.7\linewidth]{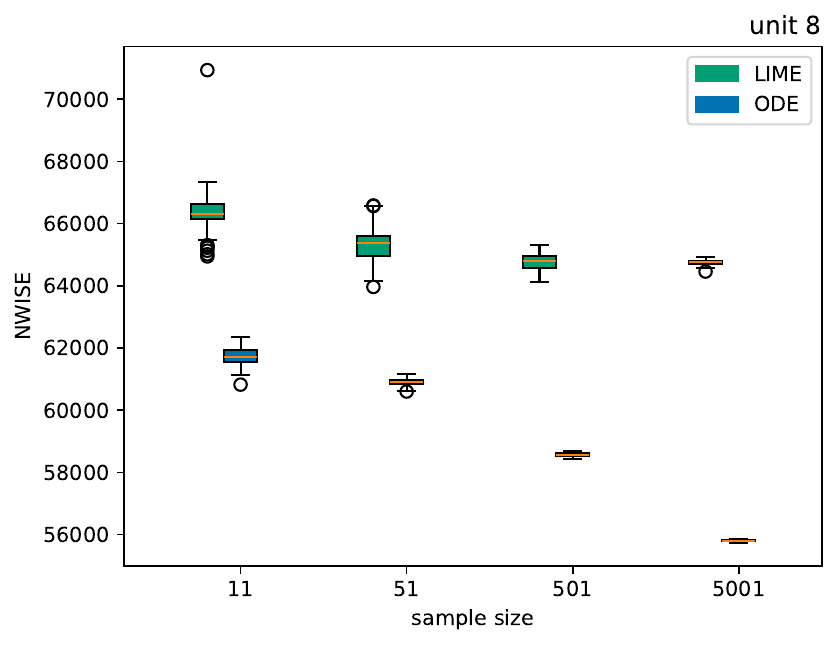}
    \caption{Boxplot of NWISE for unit 8 over different sample sizes}
    \label{fig:boxplot_NWISE_unit8}
\end{figure}

Typically, the results in LIME benefit from larger sample sizes in terms of stability. We have visualized this property by boxplots over different sample sizes with the LIME and ODE approach in Figure \ref{fig:boxplot_NWISE_unit8}. The decreasing variance in NWISE is also a property of the ODE approach as expected. When the neighborhood is specified correctly, i.e. such that it can be well approximated by a linear model, we expect this behavior to occur. For results on all units refer to Figure \ref{fig:boxplots_NWISE} in Appendix \ref{apx:boxplots-NWISE}. In some of the examples, it seems that an increase in sample size is detrimental to the NWISE. We believe that this is the result of a poorly chosen neighborhood. If the neighborhood cannot be well approximated by a linear function and more samples are drawn, the sample point corresponding to the reference point will have a smaller impact on the secondary model fit. Hence, the ODE and the LIME approach suffer from a misspecification. 

The ODE approach is also naturally beneficial for stability. The support points ``anchor'' the sampling of new points to a certain region, resulting in less variability, see e.g. Figure \ref{fig:boxplots-variability} in Appendix \ref{apx:variability-coef} for an illustration.

\section{Discussion and Outlook\label{sec:discussion}}

In this exposition we concentrated on tabular data and regression tasks but evidently design of experiments could also be applied for different data structures. It seems that in the field of XAI, image data is of particular interest. In the LIME framework, this kind of data is simplified by grouping pixels to superpixels, i.e. larger clusters of pixels, and then expressing the activation of these pixels via dummy variables. The sampling step of generating new instances works similarly in this setting, with superpixels randomly being activated. This is a setting where a classical linear regression model with locality weights as secondary model seems reasonable as well. The variables to be optimized over are then simply dummy variables of zeros and ones. 

The described method is suitable for any dimension, but becomes computationally challenging left in its pure form. One simple way to generalize this approach to higher dimensions would be to look for the design that allocates regularizing weight $\delta$ to the center point and equal weight to the other design points that are elements in the $m$-ary Cartesian product of the set $\left\{-u, u\right\}$. The formula for the D-criterion of such a design has the simple functional form
\begin{equation*}
     D_m(u, \kappa) = \frac{\delta + (1- \delta) \lambda_\kappa(\sqrt{m} u)^2}{(\delta + (1- \delta) \lambda_\kappa(\sqrt{m} u))^2 ((1-\delta) u^2)^m}
\end{equation*}
for the RBF-Kernel and Euclidean distance. This can be rewritten and optimized in one dimension much in the same way. Such a design will have a support of size $2^m + 1$. It is conjectured that smaller designs could attain equal normalized criterion values. Note also that the criterion value of the design is invariant to rotations of the design units around the center point, which can be leveraged for robustness.
In any case, there are several further, more elaborate possibilities how to efficiently cope with increasing dimensionality, which will be subject of further research.

We have not addressed feature selection in this work, although it is likely relevant in many applications. In the future, the work could be extended to settings, where there are multiple variables and only a subset of those should be included in the explanation. In LIME, feature selection is performed in different ways a priori to perturbing the data and estimating the simple secondary model. A possible extension could be to fit penalized models, e.g. lasso models that perform feature selection automatically by setting some coefficients to zero. 

Apart from the inefficiency in the sample generation in LIME, some authors have drawn attention to troubling application scenarios, where the random sampling of new data for the simple secondary model fit can cause further issues. \cite{slack2020} have raised concerns about intentional biases in (primary) classification models that can nefariously hide their true nature by detecting if an input to the prediction function originates from the true data generating process or not. In this setting, there is no direct public access to the prediction function $f$, but it is possible to provide inputs to the primary model and receive responses, thereby making post-hoc explanation possible. The idea by the providing party is to train an intermediate classification model that decides if a data point follows, what the authors call the input data's distribution, or not. If it does not, the prediction will be made by a different function that is truly unbiased, but not actually used for prediction. If the input is classified as a real input, the actual biased prediction function will be used for classification. This approach can be used to intentionally hide biases, for example, racism or sexism in ML models. 

If such a use case is expected, the natural approach would be to sample from a distribution that is as similar to the expected data as possible. One could try to improve on LIME's method by modeling the training data distribution and then sampling from this distribution or applying optimal design of experiments in such a setting. The underlying assumption is that the training data's distribution is the same as the one of data that will ultimately be used.

Another extension of this work could be the utilization of joint designs, i.e. ones that are optimal for several reference points jointly. \cite{fedorov1999} give theoretical results on such situations, and the approach could be easily adapted to the LIME framework.

This work is limited to the use of optimal design of experiments for LIME explanations, but in general, it exemplifies how optimal design of experiments can be used for efficient data collection, which results in a decrease in consumption of scarce resources like storage space and energy. We expect that by an increase in the application of machine learning in our modern society, resources will become scarce and the drive to develop efficient methods will increase. Optimal design of experiments may offer a cheap way to free up resources.

\section*{Acknowledgments}

We acknowledge the use of ChatGPT (OpenAI) to assist in editing the English grammar and improving language clarity in  the abstract and introduction of the manuscript. The content and ideas are solely the authors' own, unless cited or stated otherwise.

\printbibliography
    
\clearpage

\appendix
\section{Explanations on all Units\label{apx:explanations}}

\begin{figure}[h]
    \centering
    \includegraphics[width=0.8\linewidth]{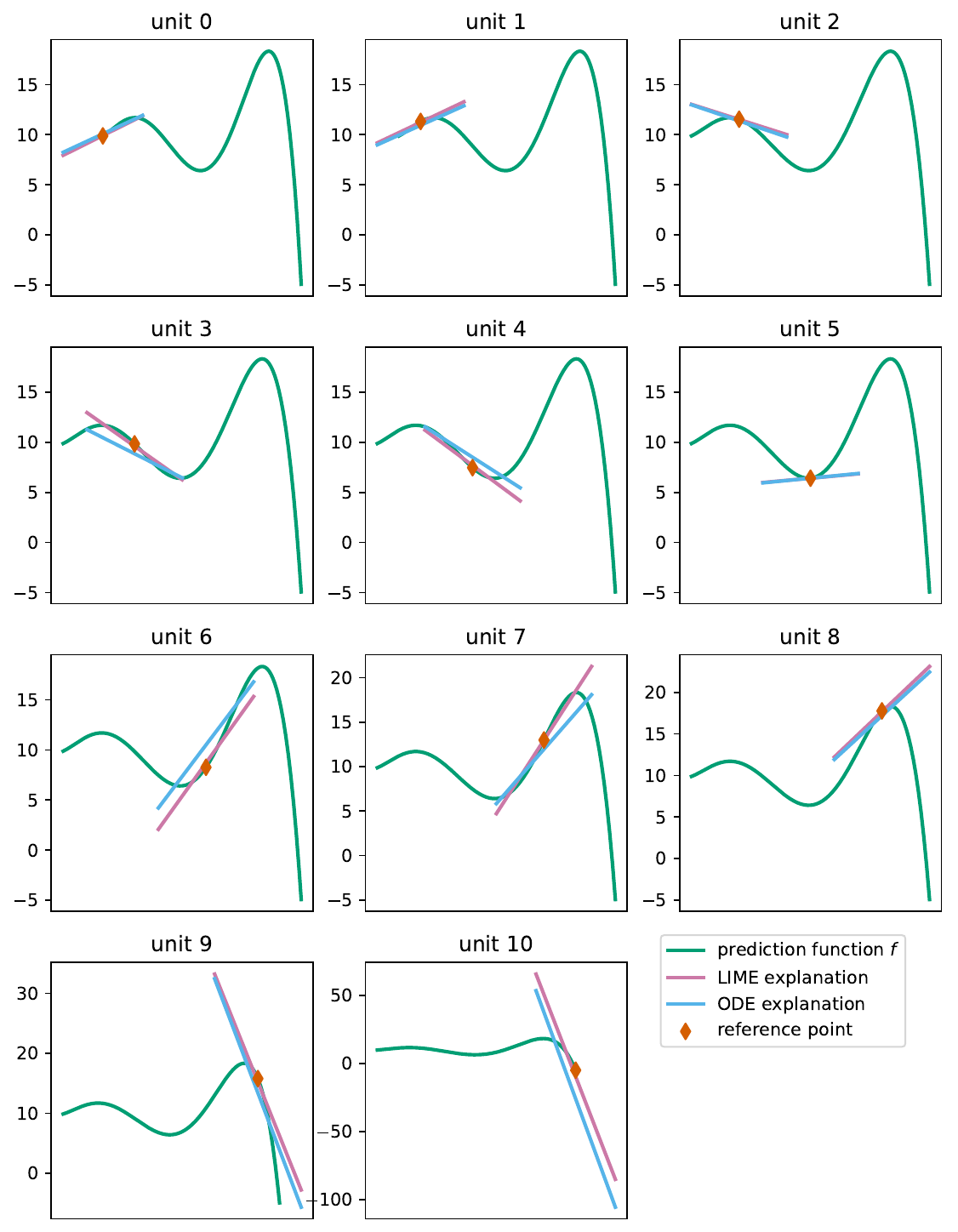}
    \caption{LIME and ODE explanations on the illustrative example}
    \label{fig:explanations-ODE-and-LIME}
\end{figure}

\clearpage

\section{Boxplots of NWISE on all Units\label{apx:boxplots-NWISE}}

\begin{figure}[h]
    \centering
    \includegraphics[width=0.8\linewidth]{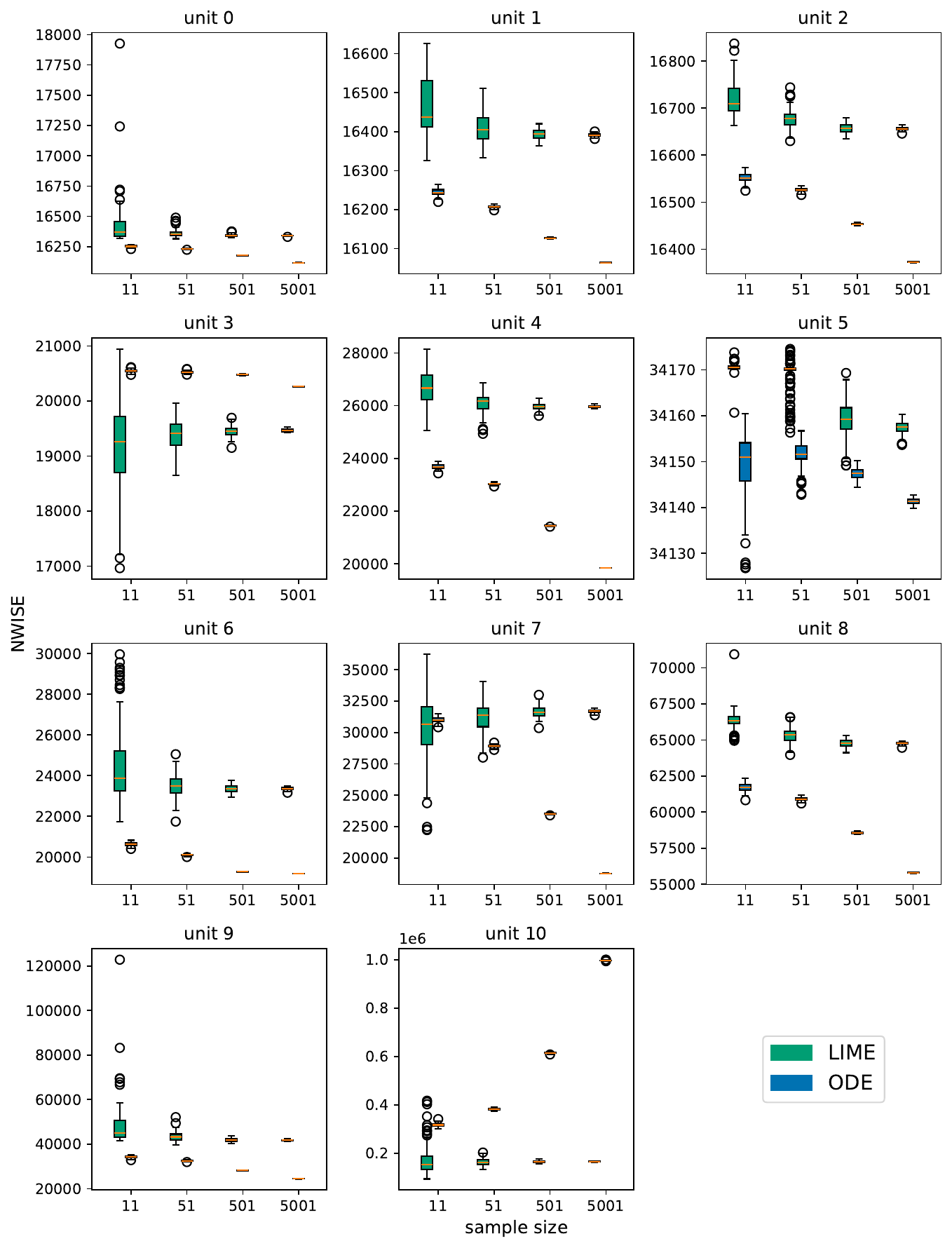}
    \caption{Boxplots of NWISE for all units over 100 runs}
    \label{fig:boxplots_NWISE}
\end{figure}

\clearpage

\section{Variability of Coefficients\label{apx:variability-coef}}

\begin{figure}[h]
    \centering
    \includegraphics[width=0.8\linewidth]{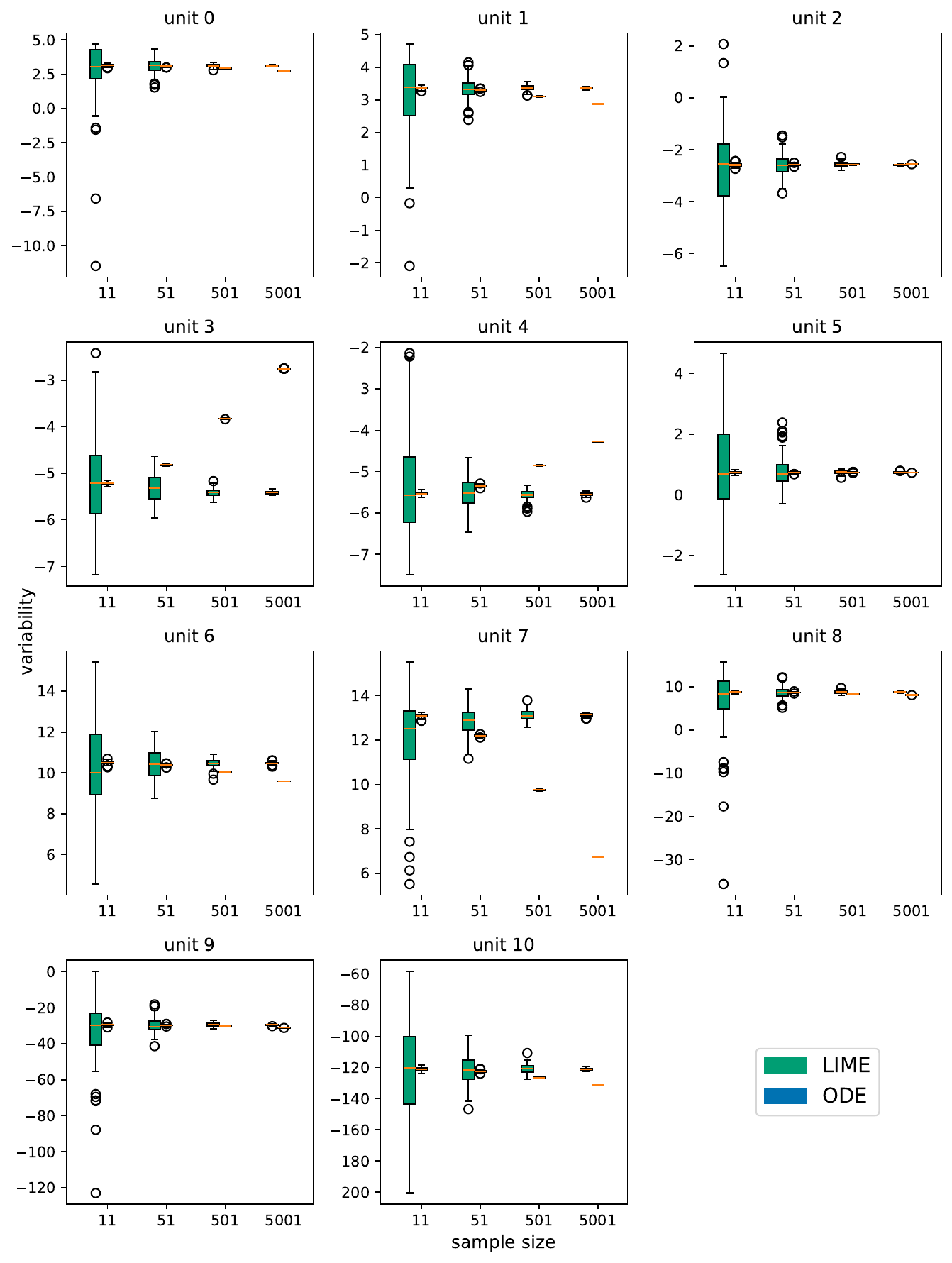}
    \caption{Variability of slope coefficients in the illustrative example}
    \label{fig:boxplots-variability}
\end{figure}

\end{document}